%% file: main.tex
\documentclass[11pt]{article}

\usepackage[final]{acl}

\usepackage{times}
\usepackage{latexsym}
\usepackage{multirow}
\usepackage{booktabs}
\usepackage[T1]{fontenc}
\usepackage{pgfplots}
\pgfplotsset{compat=1.18}
\usepackage[utf8]{inputenc}

\usepackage{microtype}
\usepackage{amssymb}

\usepackage{graphicx}
\usepackage{xspace}
\usepackage{amsmath}
\usepackage{xurl} 
\newcommand{\projname}{\textsc{RADAR}\xspace}
\title{RADAR: Rubric-Aware Dependency and Redundancy \\ Analysis for LLM-as-Judge Evaluation}

\author{
\small
Divyansh Singh$^{1,2}$ \quad
Reza Davari$^{1}$ \quad
Afra Mashhadi$^{1,3}$ \\[1pt]
\footnotesize
$^{1}$Microsoft \qquad
$^{2}$University of Florida \qquad
$^{3}$University of Washington \\[1pt]
\footnotesize
\texttt{rezadavari@microsoft.com} \qquad
\texttt{mashhadi@uw.edu} \qquad
\texttt{divyansh.singh@ufl.edu}
}

\begin{document}

\maketitle

\begin{abstract}
Rubric-based LLM-as-judge pipelines often assume that evaluation criteria provide independent signals. In practice, however, criteria can be behaviorally coupled: improving one criterion may systematically change scores on another, distorting aggregate scores used in model-release or product-update decisions. We introduce RADAR, a lightweight preflight diagnostic framework for estimating such coupling before large-scale evaluation. Given a rubric, RADAR generates targeted synthetic probes, scores each probe on all criteria, and produces a directional coupling matrix that shows which criteria co-score and how. We validate RADAR on three industry-relevant evaluation settings: NVIDIA HelpSteer2, SumPubMed, and the Yale-Salesforce SummEval benchmark. Using only a small number of probes per criterion, RADAR recovers human inter-criterion correlation structure (Pearson $r \geq 0.84$) and provides practitioners with concrete audit signals about redundancy, hierarchy, and aggregation sensitivity before committing to large-scale judging.
\end{abstract}

\section{Introduction}

LLM-as-a-judge evaluation has become a common way to scale the assessment of
open-ended model outputs \citep{zheng2023judging,chen2025judgelrm,zhang2025through}. Many pipelines now use
rubrics of criteria such as \emph{helpfulness}, \emph{correctness},
\emph{coherence}, or \emph{fluency} to decompose quality into named dimensions,
making evaluation more interpretable and actionable
\citep{liu2023geval,kim2024prometheus,ye2023flask}. In industry this scaffolding
is load-bearing: rubric scores gate model releases, drive iteration, and feed
automated regression checks \cite{shankar2024validates}, so the trustworthiness
of a rubric directly affects shipping decisions. 

\begin{figure}[!t]
  \centering
  \includegraphics[width=0.90\columnwidth]{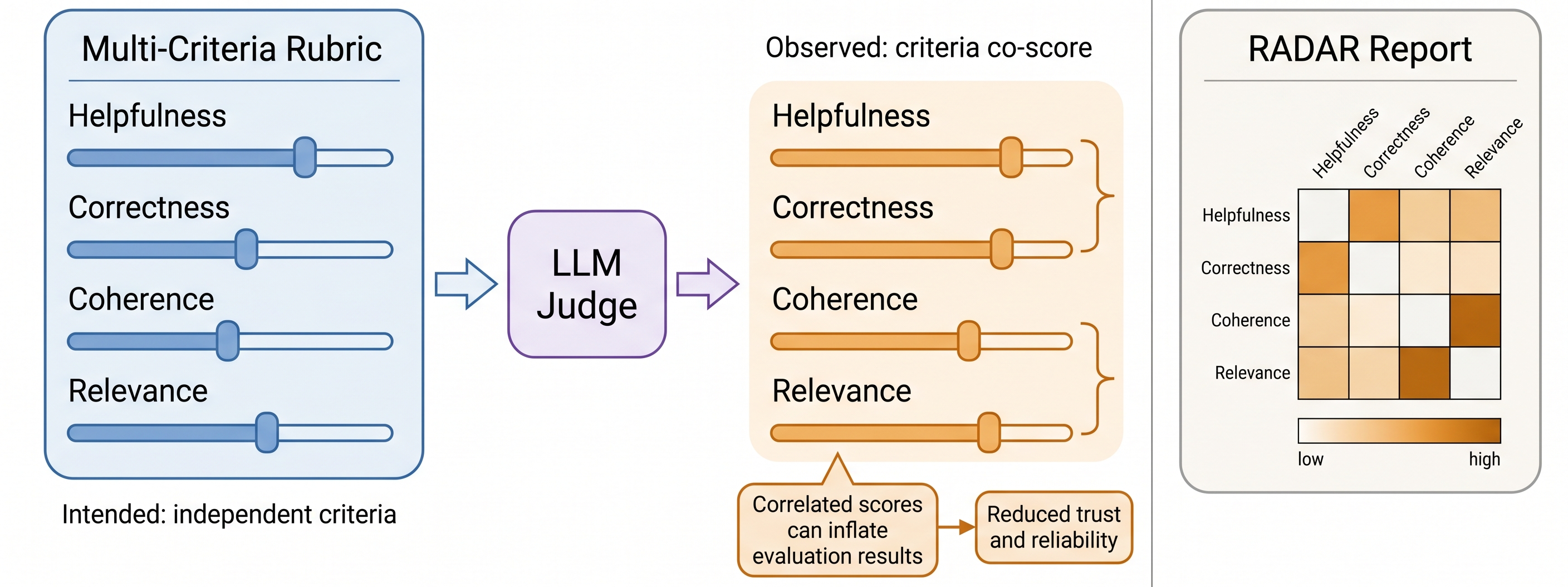}
  \caption{\small Rubric criteria can co-score under an LLM judge:
    independent-looking criteria (left) may move together (center), inflating
    aggregates. \projname{} recovers this hidden structure (right), flagging
    which criteria are coupled and how before evaluation.}
  \label{fig:cocore}
  \vspace{-0.8em}
\end{figure}

Rubric-based evaluation rests on an implicit assumption: that criteria provide
sufficiently independent signal. This can fail in two ways. The first is \emph{rubric-intrinsic}: criteria that are duplicated, nested, or
target the same latent construct, visible in the rubric text and detectable with
lightweight checks such as embedding-based criterion de-duplication or
entailment-style rubric matching
\citep{siro2026learning,dhole2026rubricraginterpretablereliablellm,shen2026rethinking}.
The second, our focus, is subtler: criteria can be semantically distinct yet
\emph{behaviorally coupled} under LLM judging \citep{murugadoss2025evaluating,
feuer2025style}. ``Conciseness'' and ``being to the point'' are not the same
criterion, yet realistic LLM outputs often make concise answers also appear
focused while verbose answers drift, so scoring both independently
double-counts the same response pattern even when no redundancy is visible in
the rubric text. Left undetected, this coupling silently inflates or distorts
aggregate scores \citep{malberg2025comprehensive}, biasing exactly the
comparisons teams rely on to choose between model versions~(Figure~\ref{fig:cocore}).

This problem is amplified by how LLM judges read rubrics. Because a judge
interprets evaluative language through its own internal sense of quality
\cite{siro2026learning}, several nominally distinct criteria can be driven by a
single property it attends to, often surface fluency, rather than by the
dimensions the rubric names. High agreement between two criteria then need not
mean the response is good on both, only that the judge read one shared feature
into both. Psychometric analyses confirm this collapse: distinct criteria reduce
to a few latent dimensions at very high inter-criterion correlations
\cite{feuer2025noise}.

Existing methods such as hierarchical criterion decomposition \cite{liu2024hdeval},
recursive rubric decomposition and reweighting \cite{shen2026rethinking}, and
pipelines built on fixed human-defined axes \cite{fabbri2021summeval,
wang2024helpsteer2opensourcedatasettraining, hashemi2025llmrubric,
siro2026learning}, largely operate on the rubric semantically or optimize it
against downstream performance, while a complementary line diagnoses judge
unreliability \emph{post hoc} on already-scored data \cite{feuer2025noise}. None
answers the preflight question: \textit{which criteria will behave as dependent
dimensions under evaluation, and in which direction, before that evaluation is
run?} Correlation in observed scores cannot separate criteria the judge treats
as one dimension from criteria that merely co-occur in the data; isolating the
former requires probing how a judge \emph{acts} on a rubric under intervention,
not just how it scores in aggregate.

We introduce \projname (Rubric-Aware Dependency and Redundancy Analysis), a
preflight framework for diagnosing LLM-as-judge
rubrics. Given a rubric,
\projname generates criterion-conditioned synthetic sample responses. That is, for each
criterion, examples intended to strongly satisfy or clearly violate it while
leaving the others unspecified. It then has a \textit{verifier} score every response on the full set of 
criteria and assembles the results into a directional coupling matrix measuring
how much changing one criterion influences the others. The question is not whether
two criteria sound similar but whether they behave as independent dimensions
under targeted intervention: if pushing \(C_i\) also raises \(C_j\), then
\(C_i\) leaks into \(C_j\). These patterns do not prescribe a unique repair,
since coupling may reflect redundancy, an intended hierarchy, or an acceptable
product preference; they do, however, convert an otherwise hidden rubric failure
mode into actionable audit hypotheses: bidirectional leakage suggests possible
double-counting, asymmetric leakage suggests upstream-downstream dependence, and
task-specific leakage suggests distribution-sensitive coupling.

We validate \projname on three human-annotated benchmarks: HelpSteer2,
SummEval, and SumPubMed, where annotation correlations provide an external
reference for criterion dependence. Across all three, \projname's coupling
estimates over axes such as helpfulness, coherence, verbosity, and relevance
predict the correlation structure observed in human annotations. A small set of
rubric-conditioned synthetic probes thus recovers the dependency structure of
much larger human-annotated corpora, indicating that preflight intervention
anticipates the coupling that surfaces in real evaluation.

Concretely, our contributions are: (i) \textbf{a criterion-coupling metric}
that, given a rubric, generates criterion-conditioned synthetic responses and
scores them on all criteria to estimate a directional leakage matrix, probing
whether criteria behave as separable dimensions under intervention rather than
by static rubric similarity; (ii) \textbf{a validation across three benchmarks}
showing that \projname{}'s synthetic coupling estimates recover human-observed
dependency structure, positioning it as a practical preflight diagnostic for
identifying co-scoring dimensions before expensive annotation or large-scale
judge evaluation; and (iii) \textbf{a low-cost scalability analysis} showing
that \projname{} recovers stable coupling estimates from only a few probes per
criterion, making it cheap enough to keep auditing the growing set of rubrics
that agentic systems, with their many sub-components, and fast-moving model
releases demand.
\section{Related Work}
Prior work motivates structured evaluation criteria but does not estimate,
before evaluation, which criteria will behave as dependent dimensions under
human or LLM judging. This is the gap \projname{} addresses.

\paragraph{Rubric-based LLM-as-a-judge evaluation}
Scoring outputs against explicit criteria is now a standard alternative to
holistic judgments, with approaches such as G-Eval \cite{liu2023geval} and
criterion-by-criterion analytic rubrics preferred for interpretability. This
decomposition makes rubric design consequential, dependence among criteria
propagates silently into the aggregate, yet whether criteria supply independent
signal is rarely checked.

\paragraph{Generating and refining rubrics}
A growing line of work uses LLMs to build and improve rubrics rather than treat
them as static: HD-Eval \cite{liu2024hdeval} learns a human-preference-guided aggregation
over a criterion hierarchy, and Recursive Rubric Decomposition (RRD)
\cite{shen2026rethinking} splits and reweights criteria, treating redundancy by
whitening and decorrelating scores. We have adapted RRD as our baseline. Both works
optimize a rubric to score well; our aim is orthogonal. We take an existing
rubric and diagnose which criteria behave as dependent dimensions, yielding not a
better score but a dependency structure for review before deployment.

\paragraph{Auditable evaluation objectives}
ARGO \cite{argo2026interpretingblackboxrewardmodels} uses rubrics to interpret black-box reward models, revealing proxy objectives that may diverge from intended behavior. \projname{} is complementary: rather than extracting rubrics from a learned reward model, it takes an existing rubric and tests whether its criteria behave as separable dimensions under LLM judging.

\paragraph{Biases and reliability of LLM judges}
Beyond known biases of length, position, phrasing, and self-preference
\cite{wang2023notfair, stureborg2024inconsistent}, \citet{siro2026learning} show
evaluative meaning is model-specific, so nominally distinct criteria can be
driven by one shared feature. \citet{feuer2025noise} document this as severe
\emph{factor collapse} (correlations above $0.9$), but measure it observationally
on already-scored data, which cannot separate criteria a judge treats as one
dimension from criteria that merely co-vary. \projname instead
intervenes, targeting one criterion while leaving others unspecified, so
movement in the untargeted criteria evidences coupling and locates where collapse
occurs before evaluation.

\paragraph{Multi-dimensional human-annotated evaluation}
Our validation uses benchmarks with human ratings along several quality axes as
an external reference for criterion dependence: SummEval \cite{fabbri2021summeval}
(coherence, consistency, fluency, relevance), HelpSteer2
\cite{wang2024helpsteer2opensourcedatasettraining} (helpfulness, correctness,
coherence, complexity, verbosity), and SumPubMed \cite{subpubmed} (coverage,
redundancy, readability, coherence, informativeness), alongside work on
human-defined rubrics for predicting human judgments \cite{hashemi2025llmrubric,
siro2026learning}. We treat their correlation structure not as a fitting target
but as an external reference for testing whether \projname's intervention-based
estimates recover dependencies that surface in real evaluation.
\section{Methodology}
\label{sec:method}
 
\begin{figure*}[t]
  \centering
  \includegraphics[width=0.66\textwidth]{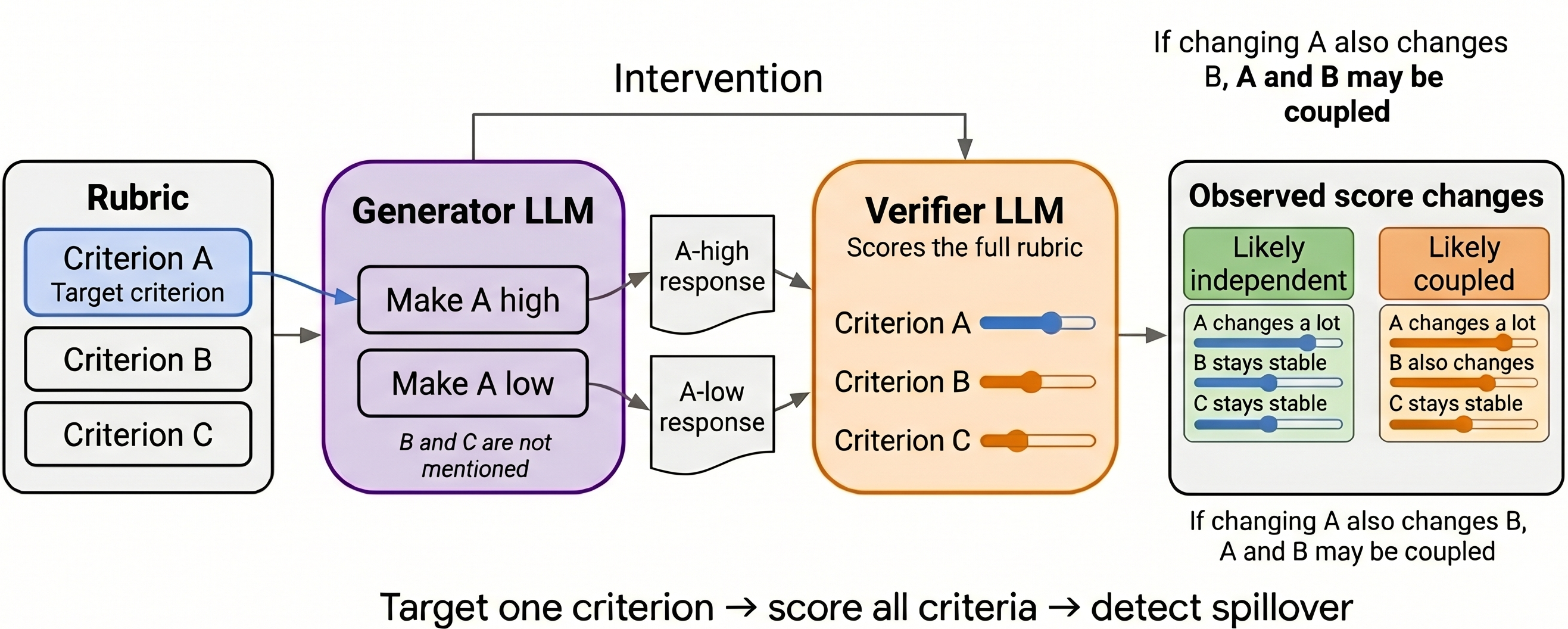}
  \caption{\small \projname{} generates high- and low-scoring probes for one criterion
and scores them on the full rubric; shifts in untargeted criteria reveal
behavioral coupling; even a few probes per criterion recover the human
inter-criterion structure (Pearson~$\geq 0.84$).}
  \label{fig:method-overview}
\end{figure*}

\projname{} takes a rubric as input and, without human-labelled data,
returns a directional coupling matrix over its criteria in three stages:
(i) criterion-conditioned \emph{synthetic intervention} on a generator
LLM, (ii) full-rubric \emph{scoring} by a verifier LLM, and (iii)
\emph{coupling statistics} from the resulting scores. Stages~(i) and~(ii)
are decoupled, so the same probes can be re-scored by different verifiers.
Figure~\ref{fig:method-overview} illustrates the design for one
target criterion.
 
\subsection{Setup and notation}
\label{sec:method:setup}
 
Let $\mathcal{R}=\{C_1,\ldots,C_n\}$ be a rubric of $n$ criteria, each a
label and a one-paragraph definition. Let $\mathcal{T}=\{t_1,\ldots,t_T\}$
be task prompts from the target domain spanning the response shapes the
rubric must cover, $G$ a \emph{generator} LLM, and $V$ a \emph{verifier}
LLM (an LLM-as-a-judge scoring responses against the rubric); both are
black boxes. For each $t$, criterion $C_i$, and direction $d\in\{+,-\}$ we
draw $N$ probe responses
\[
   x^{(t,i,d)}_{v} \sim G\bigl(\mathrm{prompt}(t,C_i,d)\bigr),
   \quad v=1,\ldots,N,
\]
a corpus of size $|\mathcal{T}|\cdot n\cdot 2\cdot N$ per generator, which
$V$ scores on \emph{every} criterion.
 
\subsection{Stage 1: Criterion-conditioned generation}
\label{sec:method:gen}
 
The prompt names only the targeted criterion $C_i$ and instructs the
generator not to optimize any other. For an \emph{evaluative} criterion
(higher~$=$~better) it asks for a response that ``strongly satisfies''
($d=+$) or ``clearly fails'' ($d=-$) $C_i$; for a \emph{descriptive} one
(e.g., verbosity), to sit at the \textsc{high} or \textsc{low} end. We
draw $N$ responses per setting at temperature $0.9$; the full template
is in Appendix~\ref{app:prompts}. Varying one criterion and measuring
shifts in the others makes this a synthetic intervention rather than
passive co-variation. Because targeting one criterion may also shift other
properties of the response, we read movement in the untargeted criteria as
\emph{evidence} of coupling, not as proof that the intervention caused it.
 
\subsection{Stage 2: Per-criterion scoring}
\label{sec:method:scoring}
 
The verifier is queried $n$ times per probe, \emph{independently} per
criterion: the $i$-th query supplies the task prompt, the response, and
$C_i$'s definition, and returns an integer score
$S_i(x)\in\{0,1,2,3,4\}$ with a brief rationale. Separate calls remove
within-prompt ordering and self-consistency effects, so residual
correlation across the $S_i(x)$ is less likely to be a prompt artifact.
Scores map to normalized values $p_i(x)=\tfrac{1}{4}S_i(x)\in[0,1]$ (a
quality level for evaluative criteria, a scale position for descriptive
ones).
 
\subsection{Stage 3: Coupling statistics}
\label{sec:method:metrics}
 
Statistics are computed per (generator, verifier, task) slice and
aggregated. Let $\mathcal{X}_{i,d}^{t}$ be the probes for task $t$
targeting $C_i$ in direction $d$, and
$\bar{p}_j(i,d) = \tfrac{1}{|\mathcal{X}_{i,d}^{t}|}
\sum_{x\in\mathcal{X}_{i,d}^{t}} p_j(x)$
the verifier's mean score on $C_j$ over that set (we suppress $t$).
 
\subsubsection{Self-effect}
The change in $C_i$'s own score between its high and low interventions,
\begin{equation}
   \mathrm{SelfEffect}(C_i) = \bar{p}_i(i,+) - \bar{p}_i(i,-) \in [-1,1],
   \label{eq:self}
\end{equation}
measures how legible the intervention is to the verifier; it is the unit
of intervention strength that all later quantities are normalized against.
 
\subsubsection{Directional coupling and leakage}
The effect of intervening on $C_i$ as read on $C_j$,
$\mathrm{GenCoupling}(C_i\!\to\! C_j) = \bar{p}_j(i,+) - \bar{p}_j(i,-)$,
normalized by the source self-effect, gives
\begin{equation}
   \mathrm{Leakage}(C_i\!\to\! C_j)
   = \frac{\mathrm{GenCoupling}(C_i\!\to\! C_j)}{\mathrm{SelfEffect}(C_i)}.
   \label{eq:leak}
\end{equation}
Diagonal entries are $1$; off-diagonals near zero mean $C_j$ is separable
from $C_i$, near one that the verifier moves them together. Leakage may be
negative (a tradeoff) or exceed one (the untargeted criterion shifts more
than the targeted one).
 
\subsubsection{Reliability gating}
Leakage is undefined at zero self-effect and unstable for weak
interventions. We mark $C_i\!\to\! C_j$ \emph{unreliable} when
$\mathrm{SelfEffect}(C_i) < \tau$ ($\tau=0.40$, a $1.6$-point gap on the
$0$ to $4$ scale; below it the self-effect is too small to normalize against);
negative self-effects
(failed interventions or mis-specified polarity) are excluded.
Pair-level summaries require both directions to pass.
 
\subsubsection{Symmetric coupling and asymmetry}
For each pair $\{C_i,C_j\}$, with $\ell_{ij}=\mathrm{Leakage}(C_i\!\to\! C_j)$,
\begin{align}
   \mathrm{SymCoupling}(C_i,C_j) &= \tfrac{1}{2}\bigl(\ell_{ij}+\ell_{ji}\bigr),\\
   \mathrm{Asymmetry}(C_i,C_j) &= \bigl|\ell_{ij}-\ell_{ji}\bigr|.
\end{align}
High symmetric coupling with low asymmetry suggests a shared-quality
cluster; high asymmetry suggests directional dependence (e.g.,
upstream-downstream) under the verifier. We also report each matrix's mean positive and maximum
off-diagonal leakage (Appendix~\ref{app:metrics}).
 
\subsection{Task and model aggregation}
\label{sec:method:task-aggregation}
 
To separate coupling intrinsic to the rubric from coupling specific to
certain response types, we run the procedure per task and aggregate:
\begin{align}
   \bar{\ell}_{ij} &= \tfrac{1}{T}\textstyle\sum_{t=1}^{T}
        \mathrm{Leakage}_t(C_i\!\to\! C_j),\\
   \sigma_{ij} &=
        \mathrm{Std}_{t}\bigl[\mathrm{Leakage}_t(C_i\!\to\! C_j)\bigr].
\end{align}
High $\bar{\ell}_{ij}$ with low $\sigma_{ij}$ indicates stable,
rubric-level coupling; high leakage on a subset of tasks indicates
distribution-sensitive coupling. Running \projname{} across
generator-verifier pairs likewise separates stable rubric-level patterns
from evaluator-specific ones.

\section{Experimental Setup}
\label{sec:setup}

We instantiate \projname{} on three multi-axis evaluation benchmarks using a
generator-verifier grid of contemporary LLMs. The benchmarks provide the
rubric axes and human ratings used as an external reference in
Section~\ref{sec:eval}; \projname{} itself consumes only rubric text during
diagnosis.

\subsection{Benchmarks and rubrics}
\label{sec:setup:data}

We evaluate on three multi-axis, human-annotated benchmarks, taking each one's
released axes and rubric descriptions verbatim.
HelpSteer2~\citep{wang2024helpsteer2opensourcedatasettraining} rates assistant
responses on \emph{helpfulness}, \emph{correctness}, \emph{coherence},
\emph{complexity}, and \emph{verbosity}; SummEval~\citep{fabbri2021summeval}
rates news summaries on \emph{coherence}, \emph{consistency}, \emph{fluency}, and
\emph{relevance}; and SumPubMed~\citep{subpubmed} rates biomedical summaries on
\emph{non-redundancy}, \emph{coherence}, \emph{readability}, and
\emph{informativeness}.\footnote{The SumPubMed rubric also defines
\emph{coverage}, but the released human annotations omit it; we validate on the
four annotated axes.} For each, we hand-curate $T=5$ short task prompts from the
target domain, independent of the annotated corpus, so validation compares
\projname{}'s synthetic estimates against the dependency structure in separately
collected human ratings.

\subsection{Models and generation protocol}
\label{sec:setup:models}

We run \projname{} over a generator-verifier grid of four LLMs: GPT-5-mini,
GPT-5.5 Reasoning, Claude Opus 4.7, and Claude Sonnet 4.6, each serving as both
generator and verifier. We exclude self-pairings ($G{=}V$) to avoid the well-documented
self-preference bias of LLM judges \cite{wataoka2025selfpreferencebiasllmasajudge}, leaving $12$ off-diagonal cells per
benchmark ($36$ total).

Per benchmark-generator pair we draw $N{=}20$ probes
per $(\text{task},\text{criterion},\text{direction})$ cell, and each verifier
scores every probe on every criterion. Generation uses temperature $0.7$ for
diversity; verifier scoring is deterministic. Full prompts, task lists,
decoding settings, and preprocessing are in Appendix~\ref{app:setup}.
\subsection{Baselines and metrics}
\label{sec:setup:baselines}

We compare against two passive estimators of criterion dependence.
\textbf{Synthetic correlation} scores unconditioned generations on every
criterion and correlates the score columns, the co-variation a rubric shows
without any intervention. \textbf{RRD} \citep{shen2026rethinking}
applies its whitening-based decorrelation to an independently sampled score set.
We score every method against the human inter-criterion correlations with
Pearson's $r$, Spearman's $\rho$, and MAE. Baseline and
metric definitions are in Appendix~\ref{app:baselines}.

\begin{table*}[!tp]
\centering
\small
\setlength{\tabcolsep}{5pt}
\begin{tabular}{llrrrrrl}
\toprule
Dataset & Pair $(C_i, C_j)$ & $\ell_{ij}$ & $\ell_{ji}$ & Sym & Asym. & Hum.\ corr. & Interpretation \\
\midrule
HelpSteer2 & helpfulness / correctness & +1.02 & +0.82 & +0.92 & 0.21 & +0.94 & bidirectional cluster \\
HelpSteer2 & correctness / coherence   & +0.22 & +0.78 & +0.50 & 0.56 & +0.47 & directional (coh.\ $\to$ corr.) \\
SummEval   & coherence   / fluency     & +0.64 & +0.51 & +0.57 & 0.16 & +0.67 & bidirectional cluster \\
SummEval   & coherence   / relevance   & +0.28 & +0.53 & +0.41 & 0.27 & +0.56 & directional (rel.\ $\to$ coh.) \\
SumPubMed  & coherence   / readability & +0.62 & +0.26 & +0.44 & 0.39 & +0.62 & directional (coh.\ $\to$ read.) \\
\bottomrule
\end{tabular}
\caption{\small Diagnostic pairs surfaced by \projname{}.\protect\footnotemark{}
\projname{}'s symmetric coupling tracks the human correlation (Hum.\ corr.) on
every pair; asymmetry separates \emph{bidirectional clusters} from
\emph{directional dependence}
($\ell_{ij}=\mathrm{Leakage}(C_i\!\to\! C_j)$, by the larger leakage).}
\label{tab:top-pairs}
\end{table*}
\footnotetext{\scriptsize Values are means across the full 4$\times$3 generator-verifier grid (12 cells per benchmark, excluding self-judging); per-cell breakdowns in Appendix~\ref{app:results}.}

\section{Results}
\label{sec:eval}

We ask two questions of \projname{}. First, does its criterion-conditioned
coupling estimates recover the dependency structure in human annotations on the
same rubric (Section~\ref{sec:eval:main})? Second, what dependencies does the
directional coupling matrix surface (Section~\ref{sec:eval:diag})?
Table~\ref{tab:main-validation} reports a single representative cell
(GPT-5.5-R~$\to$~Sonnet-4.6) so that \projname{}, synthetic correlation, and
RRD~\cite{shen2026rethinking} are scored under one generator;
Table~\ref{tab:top-pairs} aggregates across the full grid, with per-cell
breakdowns for both views in Appendix~\ref{app:results}.

\subsection{\projname{} recovers human-observed criterion dependence}
\label{sec:eval:main}

For each benchmark we compare \projname{}'s symmetric coupling
$\mathrm{SymCoupling}(C_i,C_j)$ against the human inter-criterion Pearson
correlations on the off-diagonal pairs, scoring it and the two passive baselines
on all three metrics (Table~\ref{tab:main-validation}).

\begin{table}[t]
\centering
\small
\setlength{\tabcolsep}{4pt}
\begin{tabular}{llrrr}
\toprule
Dataset & Method & Pear. & Spear. & MAE \\
\midrule
\multirow{3}{*}{HelpSteer2}
  & Synthetic corr.\  & +0.671 & +0.714 & 0.258 \\
  & RRD               & +0.570 & +0.044 & 0.318 \\
  & \projname{}       & \textbf{+0.957} & \textbf{+0.879} & \textbf{0.075} \\
\midrule
\multirow{3}{*}{SummEval}
  & Synthetic corr.\  & -0.111 & -0.500 & 0.581 \\
  & RRD               & +0.050 & +0.395 & 0.341 \\
  & \projname{}       & \textbf{+0.842} & \textbf{+0.943} & \textbf{0.273} \\
\midrule
\multirow{3}{*}{SumPubMed}
  & Synthetic corr.\  & +0.765 & +0.486 & 0.160 \\
  & RRD               & +0.064 & -0.030 & 0.386 \\
  & \projname{}       & \textbf{+0.872} & \textbf{+0.829} & \textbf{0.079} \\
\bottomrule
\end{tabular}
\caption{\small Agreement with human inter-criterion Pearson correlations
(GPT-5.5-R~$\to$~Sonnet-4.6 cell). \projname{} recovers the human structure on
all three benchmarks (Pearson~$\geq 0.84$) where the
baselines~\cite{shen2026rethinking} do not.}
\label{tab:main-validation}
\end{table}

\begin{table}[t]
\centering
\small
\setlength{\tabcolsep}{6pt}
\begin{tabular}{rrrc}
\toprule
$N$ & Cost (MTok) & \% of full & Pearson $r$ \\
\midrule
3  & 1.16 & 0.8\% & $0.953 \pm 0.011$ \\
5  & 1.94 & 1.3\% & $0.953 \pm 0.009$ \\
20 & 7.74 & 5.2\% & $0.957$ \\
\bottomrule
\end{tabular}
\caption{\small Cost vs.\ accuracy by probe budget $N$ (HelpSteer2; full
annotation $149.5$M tokens). Pearson $r$ (mean $\pm$ s.d.\ over $30$ probe
subsamples; $N{=}20$ is the full sample) is flat as cost grows: $N{=}3$ costs
under $1\%$ of full annotation at the accuracy of $N{=}20$, and all three
benchmarks stay $\geq 0.84$ at every $N$ (Appendix~\ref{app:n-ablation}).}
\label{tab:cost-acc}
\end{table}

\projname{} recovers the human structure on all three benchmarks
(Pearson $\geq 0.84$, Spearman $\geq 0.83$), the only method to do so
consistently: the passive baselines are inconsistent, with synthetic correlation at
$r{=}-0.11$ on SummEval and RRD never exceeding $\rho{=}0.40$. This holds with as few as $N{=}3$ to $5$ probes per
criterion, and cheaply: cost grows with $N$ but accuracy does not
(Table~\ref{tab:cost-acc}; full sweep in Appendix~\ref{app:n-ablation}).

\subsection{Reading the audit: clusters, asymmetry, and unreliable criteria}
\label{sec:eval:diag}
The directional leakage matrix turns each rubric into an actionable audit: it
identifies which criteria behave as dependent signals, separates symmetric
double-counting from one-way dependence, and prioritizes the pairs whose
aggregation should be inspected before large-scale judging
(Table~\ref{tab:top-pairs}; full matrices in Appendix~\ref{app:results}).
\emph{Bidirectional clusters} (high symmetric coupling, low asymmetry) mark
criteria that move as one latent axis: HelpSteer2
\emph{helpfulness}/\emph{correctness} ($\mathrm{Sym}{=}0.92$,
$\mathrm{Asym}{=}0.21$) and SummEval \emph{coherence}/\emph{fluency} ($0.57$,
$0.16$). A release gate or aggregate that independently weights both criteria
may double-count that axis and over-weight it against the rest of the rubric; practitioners should
inspect whether that is happening, for example by reviewing weights, hierarchy,
or criterion definitions. \emph{Directional dependencies}, invisible to
symmetric correlation, run one way: HelpSteer2
\emph{coherence}~$\to$~\emph{correctness} ($0.78$ vs.\ $0.22$), SummEval
\emph{relevance}~$\to$~\emph{coherence} ($0.53$ vs.\ $0.28$), and SumPubMed
\emph{coherence}~$\to$~\emph{readability} ($0.62$ vs.\ $0.26$). Here the
downstream criterion inherits the upstream's signal, so the pair is not
independent evidence; the upstream criterion is the one to scrutinize first.
A single correlation would conflate these cases;
\projname{}'s asymmetry separates them, and they warrant different review.

\subsection{Reliability and robustness}
\label{sec:eval:rel}
Self-effects clear the reliability threshold $\tau{=}0.40$ on every axis (min
$0.46$; per-benchmark means $0.64$ to $0.78$), so every leakage above comes from
a judge-legible intervention; criteria below $\tau$ are flagged \emph{unreliable}
and excluded. Full tables and bootstrap CIs are in Appendix~\ref{app:results}.
\section{Discussion}

\paragraph{Rubric auditing has to scale with model and agent churn.}
Between GPT-3.5 in Nov.\ 2022 and GPT-4o in May 2024, OpenAI released six named public text or reasoning models; between GPT-4o and GPT-5.5 in Apr.\ 2026, that number rose to 31~\citep{openai2026releasenotes}, so every release can force a fresh look at the rubrics that judge system behavior. Rubric-based LLM-as-judge scoring is the default at scale~\citep{masood2026rubric}, feeding the release gates and regression dashboards where coupled criteria double-count one latent factor and mis-rank model variants. Agentic systems compound this, fanning out into orchestrators and sub-agents~\citep{anthropic2025multiagent}, each needing its own rubric and audit.

\paragraph{\projname{} flags coupled criteria cheaply, before the rubric is used.}
\projname{} flags this coupling where the passive baselines miss it,
at a cost independent of corpus size: a preflight costs $\sim$0.4M, $\sim$1.2M,
and $\sim$1.9M tokens at $N{=}1$, $3$, and $5$ against $\sim$150M to annotate the
full HelpSteer2 corpus ($380\times$, $130\times$, and $77\times$ savings;
Appendix~\ref{app:cost}), and accuracy barely moves: pair-level Pearson stays
$\geq 0.84$ even at $N{=}1$
(Appendix~\ref{app:n-ablation}). A failure to steer is itself diagnostic: a
criterion the generator cannot move and the verifier cannot read (a low
self-effect) is too ill-defined to score reliably.

\paragraph{Using \projname{} in an industry loop.}
\projname{} runs on candidate rubrics before large-scale judging: teams set a
policy threshold, inspect coupled pairs above it, decide whether the pattern
reflects redundancy, hierarchy, or an acceptable product preference, and rerun
the diagnostic after any rubric change. This gives rubric design an explicit
audit step, replacing one-shot semantic review with a behavioral check of how
the evaluator will actually score.

\section{Conclusion}

We introduced \projname{}, a diagnostic that uses targeted synthetic probes to
estimate directional coupling between criteria before large-scale judging or
annotation. Across three benchmarks it recovers the human inter-criterion
structure (Pearson~$\geq 0.84$) at small probe budgets
(Appendix~\ref{app:n-ablation}).

\section*{Limitations}

\projname{} measures the coupling a rubric induces under a specific generator-verifier configuration, not an intrinsic property of the rubric text. This conditioning is deliberate: coupling is a behavioral property of the evaluator, so an audit tied to the deployed judge is exactly what a team needs before trusting its scores. The estimate should be refreshed when the production evaluator or target domain changes.

\projname{} provides behavioral evidence of criterion dependence, not formal causal identification. For a preflight audit, the relevant question is whether the verifier moves two criteria together under targeted probes. Some dimensions are genuinely hard to vary in isolation, as improving coherence can shift perceived readability, but directional normalization and self-effect gating control for intervention strength, so reported leakage is conditioned on judge-legible interventions rather than weak or failed probes.

Human inter-criterion correlation is an imperfect external reference: it mixes genuine dependence, annotator noise, and dataset-specific co-occurrence, and no benchmark provides a gold dependency graph. We use it only as a validation proxy; \projname{} itself requires no human inter-criterion labels and remains usable in production settings where such labels are unavailable.

Our validation spans three text-based, multi-axis settings, assistant response, news, and biomedical summarization, covering common industry uses of LLM-as-judge systems. The method itself is modality- and task-agnostic, depending only on a generator, a verifier, and a rubric; extending it to multimodal evaluation, tool-use agents, safety rubrics, or multi-turn workflows requires new probe tasks and intervention templates, not changes to the diagnostic.

Finally, \projname{} deliberately stops at diagnosis: it surfaces where criteria behave dependently and leaves the decision to practitioners, because in production the same coupling may be redundancy, an intended hierarchy, or an acceptable product preference. Prescribing a single fix would overreach; the diagnostic's role is to make that decision informed, and to make it before expensive annotation or judge runs.

\bibliography{custom}
\appendix
\section*{Appendix}
\input{appendix}
\end{document}

%% file: appendix.tex
\section{Preflight Auditing vs.\ Full-Corpus Annotation}
\label{app:cost}
Annotating a corpus against a rubric, by human raters or an LLM judge, pays a
per-item cost on every criterion, scaling as $\mathcal{O}(|D|\,K)$, whereas
\projname{} probes a small fixed task set at a cost independent of $|D|$
(\S\ref{app:n-ablation}). Table~\ref{tab:cost-radar-vs-annotation} quantifies the
gap: on HelpSteer2 ($|D|=21{,}362$, $K=5$), full LLM annotation costs
$\approx\!149.5$\,MTok against $1.94$\,MTok for an $N=5$ preflight ($77\times$
cheaper, or $380\times$ at $N=1$) and the gap widens to $23\times$ to $117\times$ on a
larger rubric ($|D|=10{,}000$, $K=8$). \projname{} is thus cheap enough to run as a
routine preflight: audit the rubric first, and merge, rewrite, split or drop strongly coupled axes
before committing to full-scale annotation.
\begin{table}[!ht]
\centering
\small
\setlength{\tabcolsep}{4pt}
\begin{tabular}{@{}lrr@{}}
\toprule
& HelpSteer2 & Hypothetical \\
\midrule
Corpus $|D|$              & 21{,}362 & 10{,}000 \\
Criteria $K$              & 5        & 8 \\
\midrule
LLM annotation (MTok)     & 149.5    & 112.0 \\
\midrule
\projname{} $N{=}1$  (MTok) & 0.39  & 0.96  \\
\projname{} $N{=}3$  (MTok) & 1.16  & 2.87  \\
\projname{} $N{=}5$  (MTok) & 1.94  & 4.78  \\
\projname{} $N{=}10$ (MTok) & 3.87  & 9.55  \\
\projname{} $N{=}20$ (MTok) & 7.74  & 19.10 \\
\bottomrule
\end{tabular}
\caption{\small Token cost (MTok, input$+$output) of LLM-annotating the full
corpus on $K$ axes versus a single \projname{} preflight cell at probe budget
$N$. Per-call medians from $\sim$20K cached calls: generator
$\approx$540\,in/200\,out; verifier $\approx$1{,}300\,in/100\,out.}
\label{tab:cost-radar-vs-annotation}
\end{table}
\section{Prompt Templates}
\label{app:prompts}
All prompts use a 0 to 4 integer Likert scale; the JSON \texttt{score}
$s\in\{0,\dots,4\}$ is rescaled to a \emph{normalized Likert score}
$s/4 \in [0,1]$ so that all axes share a common range. Reasoning models emit a
\texttt{<think>...</think>} block, which we strip before a generation is shown to
any verifier.
\subsection{Stage 1: criterion-conditioned generation}
\label{app:prompts:gen}
The generator sees \emph{only} the target criterion $C$ and a direction
$d\in\{+,-\}$; all other criteria are hidden. The same prompt is reused across the
$N$ variants, so diversity comes solely from sampling temperature.
\paragraph{Generator system message (all benchmarks).}
\begin{quote}\small\ttfamily
You are generating candidate responses for rubric testing. Follow the user
instructions precisely. Output ONLY the candidate response itself (no preamble,
no rubric commentary, no JSON, no surrounding quotes).
\end{quote}
\paragraph{Generator user message.}
\begin{quote}\small\ttfamily
Task:\\
\{task\_prompt\}\\[2pt]
Generate a response that \{target\_phrase\}.\\[2pt]
Criterion: \{criterion.label\}\\
\{criterion.description\}\\[2pt]
Important:\\
\{instructions\}
\end{quote}
\noindent The direction $d$ controls \texttt{\{target\_phrase\}} and
\texttt{\{instructions\}}:
\begin{itemize}\itemsep1pt
  \item \textbf{$d=+$, evaluative axis:} \emph{``strongly satisfies the following
  criterion''}; \textbf{$d=+$, descriptive axis} (e.g.\ HelpSteer2
  complexity/verbosity): \emph{``sits clearly at the HIGH-word end of the
  $\langle$label$\rangle$ spectrum''}. Instructions: optimize \emph{only} for this
  criterion, do not intentionally move any other criterion, keep the response
  realistic.
  \item \textbf{$d=-$, evaluative axis:} \emph{``clearly fails the following
  criterion''}; \textbf{$d=-$, descriptive axis:} \emph{``sits clearly at the
  LOW-word end''}. Instructions: make the failure realistic and plausible, do not
  modify other criteria, do not make the response nonsensical unless the criterion
  requires it.
\end{itemize}
\subsection{Stage 2: per-criterion scoring}
\label{app:prompts:judge}
Each verifier scores every probe independently, issuing one call per criterion (5
for HelpSteer2, 4 for SummEval/SumPubMed). The system message names a single
criterion and gives its one-paragraph definition; the user message supplies the
prompt/response pair and the JSON schema.
\paragraph{Verifier user message (HelpSteer2 form).}
\begin{quote}\small\ttfamily
Rate the \{criterion\_label\} of the following response on a 5-point Likert scale
(integer 0-4) for this criterion.\\[2pt]
\#\#\# Prompt\\
```\\
\{prompt\}\\
```\\[2pt]
\#\#\# Response\\
```\\
\{response\}\\
```\\[2pt]
\#\#\# Output schema\\
Return exactly this JSON object, with no surrounding text. Fill the `reasoning`
field first with a short step-by-step analysis (3-6 brief steps), then choose the
integer `score`, then write a 1-2 sentence `rationale`.\\
\{\\
\ \ "reasoning": "Step 1: ...\textbackslash nStep 2: ...",\\
\ \ "score": 0,\\
\ \ "rationale": "1-2 sentences explaining the score."\\
\}
\end{quote}
The SummEval/SumPubMed forms are identical except that
\texttt{\#\#\# Prompt}/\texttt{\#\#\# Response} become
\texttt{\#\#\# Source article}/\texttt{\#\#\# Candidate summary}.
\paragraph{Verifier system messages.}
Each criterion has its own system message of the form ``You are an expert annotator
rating only the $\langle$CRITERION$\rangle$ \dots on a 5-point Likert scale
(0 to 4)\dots'' followed by the definition. The full set of definitions is given in
Tables~\ref{tab:crit-hs2} to \ref{tab:crit-spm}.
\begin{table}[t]\centering\small
\setlength{\tabcolsep}{4pt}
\renewcommand{\arraystretch}{1.2}
\begin{tabular}{@{}p{0.29\columnwidth}p{0.59\columnwidth}@{}}
\toprule
Criterion & Definition (verbatim from system message) \\
\midrule
Helpfulness / Understanding & How useful and helpful the response is (an overall
quality rating). \\
Correctness / Completeness & Based on facts, no hallucinations or mistakes; covers
everything required in the instruction. \\
Coherence / Clarity & Self-consistent in content and style, does not contradict
itself, can be logically followed, no redundant or repeated information. \\
Language Complexity & Descriptive axis (simple\,$\rightarrow$\,complex): simple end
uses child-level vocabulary; complex end uses expert-level language. \emph{Higher is
not better.} \\
Verbosity & Descriptive axis (succinct\,$\rightarrow$\,verbose): succinct end is
direct; verbose end is wordy/long-winded. \emph{Higher is not better.} \\
\bottomrule
\end{tabular}
\caption{HelpSteer2 criterion definitions used by both the generator and the
verifier.}
\label{tab:crit-hs2}
\end{table}
\begin{table}[t]\centering\small
\setlength{\tabcolsep}{4pt}
\renewcommand{\arraystretch}{1.2}
\begin{tabular}{@{}p{0.29\columnwidth}p{0.59\columnwidth}@{}}
\toprule
Criterion & Definition \\
\midrule
Coherence & Collective quality of all sentences: well-structured and
well-organized, builds from sentence to sentence, not a heap of related
information. \\
Consistency & Factual alignment with the source; only statements entailed by the
source. Hallucinated/contradicted facts are inconsistent. \\
Fluency & Quality of individual sentences: grammatical, well-written, free of
formatting/capitalization errors and awkward phrasing. \\
Relevance & Selection of important content; include only important information,
avoid redundancy and excess detail. \\
\bottomrule
\end{tabular}
\caption{SummEval criterion definitions (0 to 4 rescaling of the original 1 to 5 scale;
a linear shift that leaves all coupling metrics unchanged).}
\label{tab:crit-se}
\end{table}
\begin{table}[t]\centering\small
\setlength{\tabcolsep}{4pt}
\renewcommand{\arraystretch}{1.2}
\begin{tabular}{@{}p{0.29\columnwidth}p{0.59\columnwidth}@{}}
\toprule
Criterion & Definition \\
\midrule
Non-redundancy & No repeated information or near-duplicate sentences; each sentence
adds new content. \\
Coherence & Well-structured and well-organized; sentences connect logically through
clear referents and transitions. \\
Readability & Easy to read for a general scientific audience: clear sentence
structure, jargon defined, free of grammatical errors. \\
Informativeness & Content-bearing information per sentence: specific results,
mechanisms, comparisons over generic filler. \\
\bottomrule
\end{tabular}
\caption{SumPubMed criterion definitions. A fifth axis (coverage) is defined in the
code but excluded, as Gupta et al.\ release human scores for only these four axes.}
\label{tab:crit-spm}
\end{table}
\section{Coupling Summary Definitions}
\label{app:metrics}
Let $\bar{s}^{+}_{C\to C'}$ and $\bar{s}^{-}_{C\to C'}$ be the mean normalized
Likert score (in $[0,1]$) that the verifier assigns to criterion $C'$ when the
generator targets $C$ in the $+$ and $-$ directions. The directional quantities are
{\small
\begin{align}
  \mathrm{SelfEffect}(C_i) &= \bar{s}^{+}_{C_i\to C_i} - \bar{s}^{-}_{C_i\to C_i},\\
  \mathrm{GenCoupling}(C_i\!\to\! C_j) &= \bar{s}^{+}_{C_i\to C_j} - \bar{s}^{-}_{C_i\to C_j},\\
  \ell_{ij} &= \frac{\mathrm{GenCoupling}(C_i\!\to\! C_j)}{\mathrm{SelfEffect}(C_i)},
\end{align}}
with the symmetric summary
$\mathrm{SymCoupling}(C_i,C_j)=\tfrac{1}{2}(\ell_{ij}+\ell_{ji})$. Since every
quantity is a \emph{difference} of mean normalized scores, the $1/4$ rescaling
cancels in $\ell_{ij}$ and affects only the readability of $\mathrm{SelfEffect}$,
not any coupling value.
\paragraph{Reliability gate.}
A source criterion $C_i$ is usable only if its self-effect is large enough that the
intervention is judge-legible. Because leakage divides by the self-effect, a small
self-effect makes the ratio unstable (and undefined at zero), so we require
$\mathrm{SelfEffect}(C_i)\geq\tau$ with $\tau=0.40$. The value is interpretable on the
normalized scale: since scores are rescaled as $s/4$ (i.e.\ $0.25$ per Likert point),
$\tau=0.40$ asks the targeted criterion to move by at least $\sim$1.6 of the $4$ Likert
points between its high and low probes, so the verifier clearly separates the two probe
sets before any leakage routed through $C_i$ is trusted. The threshold is a conservative
floor rather than a tuned hyperparameter: every axis in our study clears it comfortably
(minimum self-effect $0.46$; Table~\ref{tab:selfeffect}), so the recovered structure is
unchanged for any threshold up to $\approx 0.45$ and no annotated axis is dropped.
Otherwise $C_i$ is flagged \emph{unreliable} and its row of leakage values is excluded
from normalized diagnoses.
\paragraph{Off-diagonal summaries.}
For compact reporting we summarize each leakage matrix with off-diagonal scalars.
Because negative leakage indicates a tradeoff, we separate the two signs into a
positive coupling mass ($\mathrm{GlobalPos}$) and a tension mass
($\mathrm{GlobalTension}$):
{\small
\begin{align}
   \mathrm{GlobalPos}
     &= \operatorname{mean}_{i\neq j}\,\max\!\left(0,\ell_{ij}\right),\\
   \mathrm{GlobalTension}
     &= \operatorname{mean}_{i\neq j}\,\max\!\left(0,-\ell_{ij}\right),
\end{align}}
and report $\mathrm{MaxCoupling} = \max_{i\neq j}\,\ell_{ij}$, the rubric's
worst-case pair.
\section{Experimental Setup Details}
\label{app:setup}
All generator and verifier calls bypass the prompt cache, so every sample is
computed fresh and every reported number reflects a real, uncached inference cost.
\paragraph{Task lists.}
Each benchmark uses five fixed, self-contained tasks (no copyrighted source text) so
the metrics step can separate rubric-intrinsic from task-dependent coupling:
\begin{itemize}\itemsep1pt
  \item \textbf{HelpSteer2:} \texttt{refund} (support reply), \texttt{summarize},
  \texttt{math} (word problem), \texttt{code} (\texttt{merge\_intervals}),
  \texttt{creative} (poem).
  \item \textbf{SummEval:} short news articles on \texttt{politics},
  \texttt{science}, \texttt{business}, \texttt{sports}, and \texttt{biography},
  each with a 2 to 3 sentence summary instruction.
  \item \textbf{SumPubMed:} fabricated biomedical abstracts on
  \texttt{clinical\_trial}, \texttt{cell\_biology}, \texttt{epidemiology},
  \texttt{pharmacology}, and \texttt{surgery}, each with a 4 to 6 sentence
  summary instruction.
\end{itemize}
\paragraph{Decoding and sampling.}
Table~\ref{tab:decoding} lists the settings. Generation uses temperature $0.9$ so the
$N=20$ variants per $(\text{task},\text{criterion},\text{direction})$ cell are
genuinely distinct; verifier scoring is deterministic (temperature $0$), with one
call per criterion.
\begin{table}[t]\centering\small
\begin{tabular}{@{}lcc@{}}
\toprule
Setting & Generator & Verifier \\
\midrule
Temperature & $0.9$ & $0.0$ \\
Max new tokens & $800$ & $800$ \\
$N$ per $(t,c,d)$ cell & \multicolumn{2}{c}{$20$} \\
\bottomrule
\end{tabular}
\caption{Decoding and sampling settings for both stages.}
\label{tab:decoding}
\end{table}
\paragraph{Cost accounting (deployment view).}
For a $K$-criterion rubric, one generator-verifier cell issues
$5\!\cdot\!K\!\cdot\!2\!\cdot\!N$ generation calls and $K$ times as many verifier
calls, i.e.\ $10KN(1{+}K)$ model calls in total. At the main-text budget $N=20$ this
is $\approx\!6{,}000$ calls for a 5-criterion rubric (HelpSteer2) and
$\approx\!4{,}000$ for a 4-criterion rubric (SummEval, SumPubMed) \emph{per cell}.
Because the recovered dependency structure is stable far below $N=20$
(\S\ref{app:n-ablation}), a production audit can be run at $N\!=\!5$
($\approx\!1{,}500$ / $1{,}000$ calls) or even $N\!=\!1$ ($\approx\!300$ / $200$
calls), turning \projname{} into a cheap preflight that fits inside a normal CI
budget. We treat $N$ as the primary cost knob and characterize it in
\S\ref{app:n-ablation}.
\paragraph{Score parsing.}
The verifier returns a single JSON object per criterion; we parse the first
balanced \texttt{\{...\}} and read its integer \texttt{score}$\in[0,4]$. Every
probe contributes to every coupling estimate.
\section{Baseline Details}
\label{app:baselines}
We compare \projname{} to two passive baselines computed on the same reported
generator-verifier cell (GPT-5.5-R $\rightarrow$ Sonnet-4.6).
\paragraph{Synthetic correlation.}
This baseline measures ordinary score co-variation without intervention. For each
benchmark task, the generator produces unconditioned responses using the standard
task prompt, with no target criterion and no direction. The verifier then scores each
response on every rubric criterion. For each criterion pair $(C_i,C_j)$, we compute
the Pearson correlation between the two score columns across generated responses.
These off-diagonal correlations form the synthetic correlation vector compared
against human inter-criterion correlations in Table~\ref{tab:main-validation}.
\paragraph{RRD.}
We use Rubric Redundancy Detection (RRD) as a decorrelation baseline. RRD starts from
an independently sampled set of model responses scored on all rubric criteria,
estimates redundancy from the empirical criterion-score covariance, and applies its
whitening/decorrelation step to identify overlap among rubric axes. We convert the
resulting pair-level redundancy scores into the same off-diagonal vector format as
\projname{} and compare it to the human inter-criterion correlation vector.
\paragraph{Comparison protocol.}
For all methods, evaluation is performed at the criterion-pair level. Given the vector
of off-diagonal method scores and the matched vector of human inter-criterion Pearson
correlations, we report Pearson's $r$ (magnitude agreement), Spearman's $\rho$ (rank
agreement), and mean absolute error (MAE). The main text reports one
fixed reference cell for readability; the complete generator-verifier grid for
\projname{} is shown in Table~\ref{tab:grid}.
\section{Full Results}
\label{app:results}
\paragraph{Grid agreement.}
Table~\ref{tab:grid} reports the agreement between each cell's $\mathrm{SymCoupling}$
vector and the human inter-criterion Pearson correlations, as Pearson\,/\,Spearman,
for all $12$ generator-verifier cells per benchmark. The cell reported in the main
text (GPT-5.5-R $\rightarrow$ Sonnet-4.6) is \textbf{bold}. Agreement is high and
stable when probes come from a capable generator such as GPT-5.5-R, recovering
the human structure across verifiers on every benchmark, which is the property a
practitioner needs before trusting the audit on a new rubric.
\begin{table}[t]\centering\footnotesize
\setlength{\tabcolsep}{2.4pt}
\renewcommand{\arraystretch}{1.12}
\begin{tabular}{@{}lcccc@{}}
\toprule
\multicolumn{5}{@{}l}{\textbf{(a) HelpSteer2}\hfill Verifier $\rightarrow$} \\
\cmidrule(r){1-1}\cmidrule(l){2-5}
Generator $\downarrow$ & {\scriptsize GPT-5-mini} & {\scriptsize GPT-5.5-R}
                       & {\scriptsize Opus-4.7} & {\scriptsize Sonnet-4.6} \\
\midrule
GPT-5-mini & - & .91/.84 & .90/.95 & .92/.96 \\
GPT-5.5-R  & .96/.89 & - & .97/.90 & \textbf{.96/.88} \\
Opus-4.7   & .87/.92 & .91/.94 & - & .93/.98 \\
Sonnet-4.6 & .90/.88 & .93/.90 & .94/.94 & - \\
\midrule
\multicolumn{5}{@{}l}{\textbf{(b) SummEval}} \\
\cmidrule(r){1-1}\cmidrule(l){2-5}
Generator $\downarrow$ & {\scriptsize GPT-5-mini} & {\scriptsize GPT-5.5-R}
                       & {\scriptsize Opus-4.7} & {\scriptsize Sonnet-4.6} \\
\midrule
GPT-5-mini & - & .91/.77 & .96/.94 & .96/.94 \\
GPT-5.5-R  & .93/.94 & - & .91/.94 & \textbf{.84/.94} \\
Opus-4.7   & .76/.77 & .71/.66 & - & .81/.89 \\
Sonnet-4.6 & .71/.71 & .73/.77 & .62/.77 & - \\
\midrule
\multicolumn{5}{@{}l}{\textbf{(c) SumPubMed}} \\
\cmidrule(r){1-1}\cmidrule(l){2-5}
Generator $\downarrow$ & {\scriptsize GPT-5-mini} & {\scriptsize GPT-5.5-R}
                       & {\scriptsize Opus-4.7} & {\scriptsize Sonnet-4.6} \\
\midrule
GPT-5-mini & - & .70/.43 & .65/.43 & .75/.37 \\
GPT-5.5-R  & .83/.49 & - & .85/.49 & \textbf{.88/.83} \\
Opus-4.7   & .47/.14 & .51/.43 & - & .60/.77 \\
Sonnet-4.6 & $-$.04/.14 & .53/.37 & .33/.20 & - \\
\bottomrule
\end{tabular}
\caption{Per-cell agreement (Pearson\,/\,Spearman) between $\mathrm{SymCoupling}$ and
human inter-criterion Pearson correlations, for every generator-verifier cell in the
four-model grid. \textbf{Bold} = the cell reported in the main text.}
\label{tab:grid}
\end{table}
\paragraph{Self-effects (reported cell).}
Table~\ref{tab:selfeffect} lists $\mathrm{SelfEffect}(C_i)$ for the reported cell.
Every axis clears the reliability gate $\tau=0.40$, so all off-diagonal leakage
values below are computed from a judge-legible intervention.
\begin{table}[t]\centering\small
\renewcommand{\arraystretch}{1.05}
\begin{tabular}{@{}llr@{}}
\toprule
Dataset & Criterion & SelfEffect \\
\midrule
\multirow{5}{*}{HelpSteer2}
 & Helpfulness & $0.88$ \\
 & Correctness & $0.61$ \\
 & Coherence   & $0.66$ \\
 & Complexity  & $0.57$ \\
 & Verbosity   & $0.46$ \\
\midrule
\multirow{4}{*}{SummEval}
 & Coherence   & $0.75$ \\
 & Consistency & $0.84$ \\
 & Fluency     & $0.71$ \\
 & Relevance   & $0.59$ \\
\midrule
\multirow{4}{*}{SumPubMed}
 & Non-redundancy  & $0.52$ \\
 & Coherence       & $0.75$ \\
 & Readability     & $0.70$ \\
 & Informativeness & $0.89$ \\
\bottomrule
\end{tabular}
\caption{Self-effects for GPT-5.5-R $\rightarrow$ Sonnet-4.6. All axes exceed the
reliability gate $\tau=0.40$.}
\label{tab:selfeffect}
\end{table}
\paragraph{Per-pair coupling (top cells).}
Table~\ref{tab:allpairs} gives every off-diagonal pair for the three best
generator-verifier cells per dataset, ranked by lowest pair-level MAE between
$\mathrm{SymCoupling}$ and the human inter-criterion Pearson correlations. For each
pair we list directional leakage $\ell_{ij},\ell_{ji}$ (rows ordered as $C_i,C_j$,
so the gap between the two columns shows the asymmetry), $\mathrm{SymCoupling}$, the
asymmetry $|\ell_{ij}-\ell_{ji}|$, and the matched human correlation. The reported
cell used throughout the main text (GPT-5.5-R $\rightarrow$ Sonnet-4.6) is shown in
\textbf{bold} within each dataset block; agreement for all $12$ cells per benchmark
is in Table~\ref{tab:grid}.
\begin{table*}[p]\centering\scriptsize
\setlength{\tabcolsep}{4pt}
\renewcommand{\arraystretch}{1.0}
\begin{tabular}{@{}llrlrrrrr@{}}
\toprule
Dataset & Gen $\rightarrow$ Ver & MAE & Pair $(C_i,C_j)$ & $\ell_{ij}$ & $\ell_{ji}$ & Sym & $|\Delta|$ & Hum.\ $r$ \\
\midrule
\multirow{30}{*}{HelpSteer2} & \multirow{10}{*}{GPT-5.5-R $\rightarrow$ Opus-4.7} & \multirow{10}{*}{0.060} & helpfulness / correctness & $+1.00$ & $+0.76$ & $+0.88$ & $0.24$ & $+0.94$ \\
 &  &  & helpfulness / coherence & $+0.10$ & $+0.70$ & $+0.40$ & $0.59$ & $+0.50$ \\
 &  &  & helpfulness / complexity & $+0.22$ & $+0.13$ & $+0.17$ & $0.09$ & $+0.18$ \\
 &  &  & helpfulness / verbosity & $+0.20$ & $+0.19$ & $+0.19$ & $0.01$ & $+0.06$ \\
 &  &  & correctness / coherence & $+0.03$ & $+0.70$ & $+0.37$ & $0.67$ & $+0.46$ \\
 &  &  & correctness / complexity & $+0.11$ & $+0.16$ & $+0.13$ & $0.05$ & $+0.18$ \\
 &  &  & correctness / verbosity & $+0.06$ & $+0.19$ & $+0.12$ & $0.12$ & $+0.06$ \\
 &  &  & coherence / complexity & $+0.06$ & $+0.05$ & $+0.05$ & $0.01$ & $+0.05$ \\
 &  &  & coherence / verbosity & $-0.06$ & $-0.01$ & $-0.04$ & $0.05$ & $-0.03$ \\
 &  &  & complexity / verbosity & $+0.39$ & $+0.40$ & $+0.39$ & $0.01$ & $+0.32$ \\
\cmidrule(lr){2-9}
 & \multirow{10}{*}{GPT-5.5-R $\rightarrow$ GPT-5-mini} & \multirow{10}{*}{0.069} & helpfulness / correctness & $+1.04$ & $+0.89$ & $+0.96$ & $0.15$ & $+0.94$ \\
 &  &  & helpfulness / coherence & $+0.20$ & $+0.65$ & $+0.42$ & $0.45$ & $+0.50$ \\
 &  &  & helpfulness / complexity & $+0.27$ & $+0.09$ & $+0.18$ & $0.18$ & $+0.18$ \\
 &  &  & helpfulness / verbosity & $+0.17$ & $+0.04$ & $+0.11$ & $0.13$ & $+0.06$ \\
 &  &  & correctness / coherence & $+0.11$ & $+0.68$ & $+0.40$ & $0.57$ & $+0.46$ \\
 &  &  & correctness / complexity & $+0.10$ & $+0.12$ & $+0.11$ & $0.02$ & $+0.18$ \\
 &  &  & correctness / verbosity & $+0.09$ & $+0.16$ & $+0.13$ & $0.07$ & $+0.06$ \\
 &  &  & coherence / complexity & $-0.03$ & $+0.03$ & $-0.00$ & $0.06$ & $+0.05$ \\
 &  &  & coherence / verbosity & $-0.27$ & $+0.01$ & $-0.13$ & $0.28$ & $-0.03$ \\
 &  &  & complexity / verbosity & $+0.27$ & $+0.74$ & $+0.50$ & $0.47$ & $+0.32$ \\
\cmidrule(lr){2-9}
 & \multirow{10}{*}{\textbf{GPT-5.5-R $\rightarrow$ Sonnet-4.6}} & \multirow{10}{*}{\textbf{0.075}} & helpfulness / correctness & $+0.98$ & $+0.83$ & $+0.91$ & $0.15$ & $+0.94$ \\
 &  &  & helpfulness / coherence & $+0.59$ & $+0.68$ & $+0.64$ & $0.09$ & $+0.50$ \\
 &  &  & helpfulness / complexity & $+0.19$ & $+0.21$ & $+0.20$ & $0.02$ & $+0.18$ \\
 &  &  & helpfulness / verbosity & $+0.13$ & $+0.30$ & $+0.22$ & $0.17$ & $+0.06$ \\
 &  &  & correctness / coherence & $+0.24$ & $+0.64$ & $+0.44$ & $0.40$ & $+0.46$ \\
 &  &  & correctness / complexity & $+0.09$ & $+0.21$ & $+0.15$ & $0.12$ & $+0.18$ \\
 &  &  & correctness / verbosity & $+0.15$ & $+0.21$ & $+0.18$ & $0.06$ & $+0.06$ \\
 &  &  & coherence / complexity & $-0.00$ & $+0.05$ & $+0.02$ & $0.05$ & $+0.05$ \\
 &  &  & coherence / verbosity & $-0.21$ & $+0.01$ & $-0.10$ & $0.22$ & $-0.03$ \\
 &  &  & complexity / verbosity & $+0.38$ & $+0.53$ & $+0.45$ & $0.15$ & $+0.32$ \\
\midrule
\multirow{18}{*}{SummEval} & \multirow{6}{*}{Opus-4.7 $\rightarrow$ Sonnet-4.6} & \multirow{6}{*}{0.240} & coherence / consistency & $+0.00$ & $+0.18$ & $+0.09$ & $0.18$ & $+0.52$ \\
 &  &  & coherence / fluency & $+0.80$ & $+0.64$ & $+0.72$ & $0.17$ & $+0.67$ \\
 &  &  & coherence / relevance & $+0.55$ & $+0.91$ & $+0.73$ & $0.36$ & $+0.56$ \\
 &  &  & consistency / fluency & $+0.00$ & $+0.04$ & $+0.02$ & $0.04$ & $+0.47$ \\
 &  &  & consistency / relevance & $+0.85$ & $+0.33$ & $+0.59$ & $0.52$ & $+0.55$ \\
 &  &  & fluency / relevance & $+0.08$ & $+0.28$ & $+0.18$ & $0.20$ & $+0.49$ \\
\cmidrule(lr){2-9}
 & \multirow{6}{*}{\textbf{GPT-5.5-R $\rightarrow$ Sonnet-4.6}} & \multirow{6}{*}{\textbf{0.273}} & coherence / consistency & $+0.09$ & $+0.38$ & $+0.24$ & $0.29$ & $+0.52$ \\
 &  &  & coherence / fluency & $+0.88$ & $+0.64$ & $+0.76$ & $0.24$ & $+0.67$ \\
 &  &  & coherence / relevance & $+0.42$ & $+1.17$ & $+0.80$ & $0.75$ & $+0.56$ \\
 &  &  & consistency / fluency & $+0.00$ & $+0.05$ & $+0.03$ & $0.05$ & $+0.47$ \\
 &  &  & consistency / relevance & $+0.64$ & $+0.01$ & $+0.33$ & $0.63$ & $+0.55$ \\
 &  &  & fluency / relevance & $+0.10$ & $+0.15$ & $+0.12$ & $0.05$ & $+0.49$ \\
\cmidrule(lr){2-9}
 & \multirow{6}{*}{Sonnet-4.6 $\rightarrow$ GPT-5-mini} & \multirow{6}{*}{0.285} & coherence / consistency & $-0.08$ & $+0.14$ & $+0.03$ & $0.22$ & $+0.52$ \\
 &  &  & coherence / fluency & $+0.52$ & $+0.37$ & $+0.44$ & $0.15$ & $+0.67$ \\
 &  &  & coherence / relevance & $+0.49$ & $+0.50$ & $+0.50$ & $0.01$ & $+0.56$ \\
 &  &  & consistency / fluency & $+0.02$ & $-0.02$ & $+0.00$ & $0.04$ & $+0.47$ \\
 &  &  & consistency / relevance & $+0.87$ & $+0.17$ & $+0.52$ & $0.70$ & $+0.55$ \\
 &  &  & fluency / relevance & $+0.04$ & $+0.07$ & $+0.05$ & $0.03$ & $+0.49$ \\
\midrule
\multirow{18}{*}{SumPubMed} & \multirow{6}{*}{\textbf{GPT-5.5-R $\rightarrow$ Sonnet-4.6}} & \multirow{6}{*}{\textbf{0.079}} & non-redundancy / coherence & $+0.42$ & $+0.30$ & $+0.36$ & $0.12$ & $+0.33$ \\
 &  &  & non-redundancy / readability & $+0.17$ & $+0.00$ & $+0.09$ & $0.17$ & $+0.29$ \\
 &  &  & non-redundancy / informativeness & $+0.35$ & $+0.10$ & $+0.23$ & $0.25$ & $+0.21$ \\
 &  &  & coherence / readability & $+0.88$ & $+0.50$ & $+0.69$ & $0.38$ & $+0.62$ \\
 &  &  & coherence / informativeness & $+0.32$ & $+0.57$ & $+0.45$ & $0.24$ & $+0.33$ \\
 &  &  & readability / informativeness & $+0.00$ & $+0.30$ & $+0.15$ & $0.30$ & $+0.12$ \\
\cmidrule(lr){2-9}
 & \multirow{6}{*}{Opus-4.7 $\rightarrow$ GPT-5-mini} & \multirow{6}{*}{0.133} & non-redundancy / coherence & $+0.23$ & $+0.65$ & $+0.44$ & $0.42$ & $+0.33$ \\
 &  &  & non-redundancy / readability & $+0.18$ & $+0.35$ & $+0.26$ & $0.17$ & $+0.29$ \\
 &  &  & non-redundancy / informativeness & $+0.04$ & $+0.91$ & $+0.48$ & $0.87$ & $+0.21$ \\
 &  &  & coherence / readability & $+0.76$ & $+0.04$ & $+0.40$ & $0.73$ & $+0.62$ \\
 &  &  & coherence / informativeness & $+0.04$ & $+0.45$ & $+0.24$ & $0.41$ & $+0.33$ \\
 &  &  & readability / informativeness & $+0.00$ & $+0.06$ & $+0.03$ & $0.06$ & $+0.12$ \\
\cmidrule(lr){2-9}
 & \multirow{6}{*}{GPT-5-mini $\rightarrow$ Sonnet-4.6} & \multirow{6}{*}{0.139} & non-redundancy / coherence & $+0.46$ & $+0.46$ & $+0.46$ & $0.01$ & $+0.33$ \\
 &  &  & non-redundancy / readability & $+0.38$ & $+0.84$ & $+0.61$ & $0.46$ & $+0.29$ \\
 &  &  & non-redundancy / informativeness & $+0.44$ & $+0.07$ & $+0.26$ & $0.37$ & $+0.21$ \\
 &  &  & coherence / readability & $+0.76$ & $+0.67$ & $+0.71$ & $0.10$ & $+0.62$ \\
 &  &  & coherence / informativeness & $+0.40$ & $+0.03$ & $+0.21$ & $0.37$ & $+0.33$ \\
 &  &  & readability / informativeness & $+0.54$ & $-0.06$ & $+0.24$ & $0.60$ & $+0.12$ \\
\bottomrule
\end{tabular}
\caption{Pair-level leakage for the top-3 generator-verifier cells
per dataset, selected by lowest pair-level MAE between $\mathrm{SymCoupling}$
and human inter-criterion Pearson correlations.
$\ell_{ij}=\mathrm{Leakage}(C_i\!\to\! C_j)$; $|\Delta|=|\ell_{ij}-\ell_{ji}|$
is the directional asymmetry; Hum.\ $r$ is the matched human Pearson
correlation. The reported cell used throughout the main text
(GPT-5.5-R $\rightarrow$ Sonnet-4.6) is in \textbf{bold}. Diagonal leakage
is $1.00$ by construction and omitted.}
\label{tab:topcells-pairs}\label{tab:allpairs}
\end{table*}
\paragraph{Global summaries and bootstrap CIs.}
Table~\ref{tab:app-globals} reports the three global summaries for the reported cell
on each benchmark, each with a $95\%$ confidence interval obtained by resampling
probes $500$ times, stratified by target criterion and perturbation direction. Across
all three datasets the GlobalPositive coupling is tightly estimated
($0.32$ to $0.38$, half-widths $\le 0.02$) and the GlobalTension is effectively zero,
confirming that the off-diagonal structure is dominated by positive cross-criterion
leakage rather than sampling noise. The MaxCoupling intervals
($[0.96,1.01]$, $[1.11,1.23]$, $[0.85,0.92]$) are likewise narrow, so the strongest
leakage pair on each benchmark is well separated from zero.
\begin{table}[t]\centering\small
\setlength{\tabcolsep}{4pt}
\renewcommand{\arraystretch}{1.0}
\begin{tabular}{@{}lccc@{}}
\toprule
\textbf{Dataset} & \textbf{GlobalPos.} & \textbf{Tension} & \textbf{MaxCoup.} \\
\midrule
HelpSteer2 & \shortstack{$0.32$\\[1pt]{\scriptsize$[0.31,0.34]$}}
           & \shortstack{$0.01$\\[1pt]{\scriptsize$[0.01,0.01]$}}
           & \shortstack{$0.98$\\[1pt]{\scriptsize$[0.96,1.01]$}} \\[5pt]
SummEval   & \shortstack{$0.38$\\[1pt]{\scriptsize$[0.36,0.39]$}}
           & \shortstack{$0.00$\\[1pt]{\scriptsize$[0.00,0.00]$}}
           & \shortstack{$1.17$\\[1pt]{\scriptsize$[1.11,1.23]$}} \\[5pt]
SumPubMed  & \shortstack{$0.33$\\[1pt]{\scriptsize$[0.32,0.35]$}}
           & \shortstack{$0.00$\\[1pt]{\scriptsize$[0.00,0.01]$}}
           & \shortstack{$0.88$\\[1pt]{\scriptsize$[0.85,0.92]$}} \\
\bottomrule
\end{tabular}
\caption{Global coupling summaries for the reported cell (GPT-5.5-R generator,
Sonnet-4.6 verifier), with $95\%$ bootstrap confidence intervals ($500$ resamples).}
\label{tab:app-globals}
\end{table}
\section{Probe Budget and Deployment Cost}
\label{app:n-ablation}
The single most important property for industrial use is that \projname{} stays
faithful at a small probe budget: $N$ is the dominant cost driver
(\S\ref{app:setup}), so the question of how small $N$ can be set without distorting
the recovered dependency structure determines whether the audit is cheap enough to
run routinely. We vary the number of probe responses $N$ per
$(\text{task},\text{criterion},\text{direction})$ cell and measure how agreement with
human inter-criterion correlations degrades as $N$ shrinks. Crucially, we subsample
the existing $N=20$ probes \emph{without rerunning any generation or scoring}: for
each $N \in \{1,2,3,4,5,10\}$ we draw $30$ random subsets stratified by
(task, target criterion, direction) and report the mean and standard deviation of
pair-level Pearson, Spearman, and MAE against the human correlation vector. $N=20$
uses all probes (no resampling).
\paragraph{Setup.}
Reported cell: GPT-5.5-R generator $\rightarrow$ Sonnet-4.6 verifier (the main-text
setting). HelpSteer2 has $10$ criterion pairs; SumPubMed has $6$. SummEval is excluded
from this ablation because probe-level scores were not retained for the reported cell
(only aggregate metrics).
\begin{table}[!ht]\centering\small
\setlength{\tabcolsep}{5pt}
\renewcommand{\arraystretch}{1.1}
\begin{tabular}{@{}rccc@{}}
\toprule
$N$ & Pearson & Spearman & MAE \\
\midrule
\multicolumn{4}{@{}l}{\textbf{HelpSteer2} (10 pairs)} \\
\midrule
1  & $+0.940 \pm 0.025$ & $+0.904 \pm 0.043$ & $0.088 \pm 0.018$ \\
2  & $+0.947 \pm 0.016$ & $+0.900 \pm 0.037$ & $0.084 \pm 0.013$ \\
3  & $+0.953 \pm 0.011$ & $+0.893 \pm 0.017$ & $0.078 \pm 0.009$ \\
4  & $+0.953 \pm 0.010$ & $+0.895 \pm 0.025$ & $0.077 \pm 0.009$ \\
5  & $+0.953 \pm 0.009$ & $+0.894 \pm 0.027$ & $0.077 \pm 0.008$ \\
10 & $+0.956 \pm 0.005$ & $+0.886 \pm 0.014$ & $0.075 \pm 0.005$ \\
20 & $+0.957$           & $+0.879$           & $0.075$           \\
\midrule
\multicolumn{4}{@{}l}{\textbf{SummEval}$^{\dagger}$ (6 pairs)} \\
\midrule
1  & $+0.846 \pm 0.029$ & $+0.977 \pm 0.035$ & $0.273 \pm 0.015$ \\
2  & $+0.830 \pm 0.025$ & $+0.962 \pm 0.027$ & $0.272 \pm 0.011$ \\
3  & $+0.841 \pm 0.020$ & $+0.964 \pm 0.028$ & $0.273 \pm 0.009$ \\
4  & $+0.839 \pm 0.013$ & $+0.958 \pm 0.025$ & $0.275 \pm 0.006$ \\
5  & $+0.840 \pm 0.011$ & $+0.960 \pm 0.026$ & $0.274 \pm 0.008$ \\
10 & $+0.841 \pm 0.008$ & $+0.954 \pm 0.023$ & $0.275 \pm 0.002$ \\
20 & $+0.842$           & $+0.943$           & $0.273$           \\
\midrule
\multicolumn{4}{@{}l}{\textbf{SumPubMed} (6 pairs)} \\
\midrule
1  & $+0.872 \pm 0.052$ & $+0.846 \pm 0.081$ & $0.093 \pm 0.025$ \\
2  & $+0.869 \pm 0.033$ & $+0.834 \pm 0.043$ & $0.088 \pm 0.014$ \\
3  & $+0.859 \pm 0.034$ & $+0.834 \pm 0.031$ & $0.087 \pm 0.014$ \\
4  & $+0.871 \pm 0.027$ & $+0.838 \pm 0.044$ & $0.083 \pm 0.013$ \\
5  & $+0.876 \pm 0.022$ & $+0.829 \pm 0.026$ & $0.082 \pm 0.008$ \\
10 & $+0.876 \pm 0.011$ & $+0.829 \pm 0.000$ & $0.080 \pm 0.004$ \\
20 & $+0.876$           & $+0.829$           & $0.079$           \\
\bottomrule
\end{tabular}
\caption{Agreement between \projname{} ($\mathrm{SymCoupling}$) and human
inter-criterion Pearson correlations as a function of probe budget $N$ per
(task, criterion, direction) cell. Means and standard deviations are over $30$
stratified subsamples of the $N=20$ probes; the $N=20$ row is the unique full sample.
Reported cell: GPT-5.5-R $\rightarrow$ Sonnet-4.6.}
\label{tab:n-ablation}
\end{table}
\begin{figure}[!ht]\centering
\begin{tikzpicture}
\begin{axis}[
  width=\columnwidth, height=5.0cm,
  xlabel={Probe budget $N$ per cell},
  ylabel={Pearson $r$ vs.\ human},
  xmode=log, log basis x=10,
  xtick={1,2,3,5,10,20}, xticklabels={1,2,3,5,10,20},
  xmin=0.85, xmax=23,
  ymin=0.80, ymax=1.0,
  ytick={0.80,0.85,0.90,0.95,1.0},
  tick label style={font=\scriptsize},
  label style={font=\footnotesize},
  legend style={font=\scriptsize, at={(0.98,0.05)}, anchor=south east,
                draw=none, fill=none},
  grid=both, grid style={gray!18, line width=0.3pt},
  every axis plot/.append style={line width=1pt},
]
\addplot[gray, dashed, line width=0.6pt, forget plot] coordinates {(0.85,0.957)(23,0.957)};
\addplot[gray, dashed, line width=0.6pt, forget plot] coordinates {(0.85,0.876)(23,0.876)};
\addplot+[mark=*, mark size=1.6pt,
  error bars/.cd, y dir=both, y explicit, error bar style={line width=0.6pt}]
  coordinates {
    (1,0.940)+-(0,0.025) (2,0.947)+-(0,0.016) (3,0.953)+-(0,0.011)
    (4,0.953)+-(0,0.010) (5,0.953)+-(0,0.009) (10,0.956)+-(0,0.005)
    (20,0.957)+-(0,0.0)
  };
\addlegendentry{HelpSteer2}
\addplot+[mark=square*, mark size=1.5pt,
  error bars/.cd, y dir=both, y explicit, error bar style={line width=0.6pt}]
  coordinates {
    (1,0.872)+-(0,0.052) (2,0.869)+-(0,0.033) (3,0.859)+-(0,0.034)
    (4,0.871)+-(0,0.027) (5,0.876)+-(0,0.022) (10,0.876)+-(0,0.011)
    (20,0.876)+-(0,0.0)
  };
\addlegendentry{SumPubMed}
\end{axis}
\end{tikzpicture}
\caption{Probe-budget sensitivity. Pair-level Pearson agreement with human
inter-criterion correlations as a function of $N$ (log scale), with $\pm1$ s.d.\ over
$30$ stratified subsamples; dashed lines mark the full-budget ($N{=}20$) value for
each benchmark. Agreement is essentially flat from $N{=}1$ onward, so the bulk of the
$20\times$ probe budget buys variance reduction, not fidelity.}
\label{fig:n-ablation}
\end{figure}
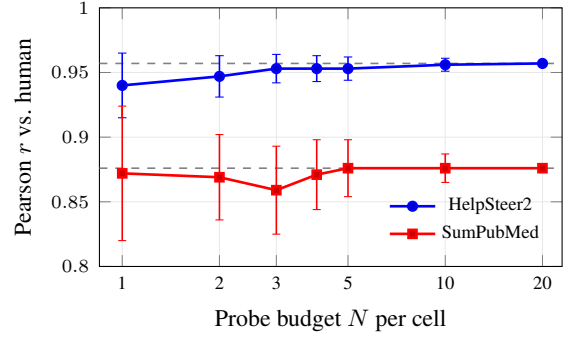
\paragraph{Takeaways.}
(i)~\textbf{Agreement saturates almost immediately.} By $N=3$, Pearson and MAE are
within $\sim$0.5\% of their $N=20$ values on HelpSteer2 and within $\sim$1\% on
SumPubMed (Table~\ref{tab:n-ablation}); the extra probes
buy lower variance, not a different answer.
(ii)~\textbf{Even a single probe per cell is informative.} At $N=1$, Pearson already
exceeds $+0.87$ on both benchmarks, with the central tendency close to the full-budget
value and only modestly higher variance ($\sigma \le 0.05$).
(iii)~\textbf{Low $N$ is the recommended operating point.} For practitioners this
means \projname{} can be run as a cheap preflight at $N\in[3,5]$ (a
$4{\times}$ to $7{\times}$ reduction in probe cost relative to our main runs, i.e.\ on
the order of $1{,}000$ to $1{,}500$ model calls per generator-verifier cell) without
materially changing the recovered dependency structure. We retain $N=20$ in the main
text for low-variance reporting; for deployment we recommend defaulting to $N=5$
(or $N=3$ when budget is tight), and reserving larger $N$ only when tight confidence
intervals on individual pairs are required.